\documentclass[conference]{IEEEtran}

\usepackage{balance}

\IEEEoverridecommandlockouts 
\usepackage{graphicx}
\usepackage{color}
\usepackage[utf8]{inputenc}
\usepackage{algorithmic}
\usepackage{algorithm}
\usepackage{bbm}
\usepackage{listings}
\usepackage[OT4,T1]{fontenc}
\usepackage[cmex10]{amsmath}
\usepackage{url}
\usepackage{multirow}

\newcommand{\specialcell}[2][c]{%
  \begin{tabular}[#1]{@{}c@{}}#2\end{tabular}}

\newcommand{\Name}[2]{\textbf{#1}$_\mathbf{#2}$}

\title{Early Warning System for Seismic Events\\ in Coal Mines Using Machine Learning}

\author{
\IEEEauthorblockN{Robert Bogucki, Jan Lasek, Jan Kanty Milczek, Michał Tadeusiak}
\IEEEauthorblockA{deepsense.io,\\
Email: \{robert.bogucki, jan.lasek, jan.milczek, michal.tadeusiak\}@deepsense.io}
}

\begin{document}
\maketitle              

\begin{abstract}

This document describes an~approach to the~problem of predicting dangerous seismic events
in active coal mines up to 8 hours in advance.
It was developed as a~part of the
AAIA`16 Data Mining Challenge: Predicting Dangerous Seismic Events in Active Coal Mines.
The~solutions presented consist of ensembles of various predictive models
trained on different sets of features. The best one achieved a~winning score of 0.939 AUC.

\end{abstract}

\section{Introduction}
\IEEEoverridecommandlockouts\IEEEPARstart{I}{n} 2015, the mining industry in Poland reported 2158 dangerous incidents
with 19 casualties and 12 severe injuries~\cite{WUG2015}. Underground mining work 
poses a~number of threats including fires, methane outbreaks or seismic tremors and bumps. 
Monitoring and decision support systems might play an~essential role in limiting the~number of incidents and their prevention.
Such systems, often based on machine learning or data mining techniques, can be effectively applied
to lessen the danger to employees and prevent potential losses arising from lost and damaged equipment, 
see, e.g.,~\cite{Zagorecki2015, Janusz2016, sikora2011induction}.

In this paper, we present a~model for predicting dangerous seismic events in coal mines. Using different 
machine learning models, we address the~classification problem of whether the~total seismic energy in the upcoming few hours
is going to reach a~warning level.  
The model was developed for the \textit{AAIA`16 Data Mining Challenge: Predicting Dangerous Seismic Events in Active Coal Mines} 
and proved  to be the~most successful approach among the 203 teams participating in the~challenge~\cite{Janusz2016}. 

The paper is structured as follows: the first section outlines the problem and describes the main challenges. Next, we describe 
our approach, focusing on feature engineering, model optimization and evaluation. Finally, in~the~last section we conclude the~work.

\section{The challenge}
In this section we introduce the~problem
and describe the~provided data. We also make some preliminary
remarks about the~data and its nature.

\subsection{Problem statement}

The given problem is a~classification task. The~goal is to develop a~prediction model that, based 
on the~recordings from 
a 24-hour long period, predicts whether an~energy warning level is going to be reached in the~upcoming 8 hours.
The warning is reached when the~total energy of seismic bumps exceeds 50~000 J = 50 kJ.
The accuracy of a~model is determined with respect to the Area Under ROC curve (AUC) metric. This 
accuracy measure is defined as follows.
Let $(\mathbf{x}_i, y_i) \in \mathbf{X}$
denote an~instance from the dataset $\mathbf{X}$,
i.e., $\mathbf{x}_i$ 
stands for the feature vector associated with a single measurement and $y_i \in \{0,1\}$ stands for its label. 
Let $f$ be a~model that maps each instance to probability that it belongs to class `1' (or, more generally, a~real-valued risk score). Then AUC is derived as
\begin{equation}\label{AUC}
AUC(f, \mathbf{X}) = \frac{\sum_{i: y_i = 0} \sum_{j: y_j = 1} \mathbbm{1}(f(\mathbf{x}_i) < f(\mathbf{x}_j))}
{|\{y_i : y_i = 0\}| \cdot |\{y_j : y_j = 1\}|}
\end{equation}
where $\mathbbm{1}(\cdot)$ denotes an indicator function that returns 1 if a~given condition is satisfied or 0 otherwise, and $|S|$ denotes the cardinality of set $S$. 
This accuracy measure returns values in the range range $[0,1]$, where 1 is achieved by a~perfect predictor. 
A random predictor yields values around $0.5$.


\subsection{Data}
Two sets of observations were provided: training dataset with accompanying labels and the test set without them. The~former was provided so that the problem could be approached from a machine learning angle, 
the~latter serves for evaluation purposes. 
The competitors were asked to submit the~likelihood of the~label `warning' for each record in the~test set.

In total, the~training set consists of 133\:151 observations. Each observation (instance) is described by a~set of 541 numbers.
Below, we briefly introduce the~data provided. For a~more thorough description of the dataset 
please refer to the~competition website~\cite{KnowledgePit2016}.

The instances are described by a set of 13 features of different type and 22 time series over last 24 hours prior to the forecasting period. 
The time series' names are followed by 1, 2, $\dots$, 24, indicating consecutive hours
of measurements (with the~most recent hour prior to forecasting period being 24). Possible suffixes 
\texttt{\_e$\xi$}, $\xi \in \{\texttt{2},\texttt{3},\texttt{4},\texttt{5},\texttt{6plus}\}$ refer to orders of magnitude of a~given time series in a certain range, e.g. \texttt{sum\_e3.5} 
stands for sum of energies within range $(10^2, 10^3]$ in the~5th hour of the~time series. 
The series are listed below:

\begin{itemize}
    \item \texttt{count\_e2}, $\dots$, \texttt{count\_e6plus} - number of registered seismic bumps;
    \item \texttt{sum\_e2}, $\dots$, \texttt{sum\_e6plus} - sum of energy of registered seismic bumps;
    \item \texttt{total\_number\_of\_bumps};
    \item \texttt{number\_of\_rock\_bursts};
    \item \texttt{number\_of\_destressing\_blasts};
    \item \texttt{highest\_bump\_energy}.
\end{itemize}

\noindent
Additionally, for the~most active geophones, the following series are provided:

\begin{itemize}
    \item \texttt{max\_gactivity};
    \item \texttt{max\_genergy};
    \item \texttt{avg\_gactivity};
    \item \texttt{avg\_genergy}; 
    \item \texttt{max\_difference\_in\_gactivity};
    \item \texttt{max\_difference\_in\_genergy};
    \item \texttt{avg\_difference\_in\_gactivity};
    \item \texttt{avg\_difference\_in\_genergy}.
\end{itemize}

\noindent
There are also 4 \textit{assessments} provided by experts. They are provided as categorical variables with four levels 
ranging from `$a$' (the lowest risk) to `$d$' (the highest risk):

\begin{itemize}
    \item \texttt{latest\_seismic\_assessment};
    \item \texttt{latest\_seismoacoustic\_assessment};
    \item \texttt{latest\_comprehensive\_assessment}; 
    \item \texttt{latest\_hazards\_assessment}.
\end{itemize}

\noindent
Finally, several features which we will refer to as \textit{general} are provided:

\begin{itemize}
    \item \texttt{total\_bumps\_energy};
    \item \texttt{total\_tremors\_energy};
    \item \texttt{total\_destressing\_blasts\_energy};
    \item \texttt{total\_seismic\_energy};
    \item \texttt{latest\_progress\_estimation\_l};
    \item \texttt{latest\_progress\_estimation\_r};
    \item \texttt{latest\_maximum\_yield};
    \item \texttt{latest\_maximum\_meter}.
\end{itemize}

\noindent \textbf{Metadata}:
For each observation we are given its location, i.e., a~\textit{longwall} in a~particular coal mine 
that the~measurement comes from. Each location is accompanied with additional information (metadata included
in a~separate file):
\begin{itemize}
    \item \texttt{main\_working\_id} - ID of the~main working site (at a~longwall);
    \item \texttt{main\_working\_name} - name of the~main working site;
    \item \texttt{region\_name} - name of region where the~main working site is located;
    \item \texttt{bed\_name} - name of coal bed;
    \item \texttt{main\_working\_type} - type of the~main working site;
    \item \texttt{main\_working\_height} - height of the~main working site;
    \item \texttt{geological\_assessment} - geological assessment of the~main working site made by experts before the~beginning of exploration (ordered categorical variable ranging from `$a$' (lower risk) to `$c$' (higher risk)).
\end{itemize}
Most of metadata were unique to the~working sites, therefore were discarded early due to the reasons discussed later.
The~only information of potential use were \texttt{main\_working\_height} and \texttt{geological\_assessment}, 
however they still had to be treated with caution:

\begin{itemize}
 \item \texttt{geological\_assessment}: A~closer insight revealed that there is none mine assessed as `$d$' and only one marked as `$c$'. It was replaced by `$b$'. Moreover, the~proportion of `$a$' assessments for longwalls in the~training and test dataset varied significantly, 25\% to 48\%, respectively.
 \item \texttt{main\_working\_height}: many working sites had unique working heights - this posed a~danger that the~feature would be used by a model as a proxy for particular location rather than a~potentially valuable information about the~height. One solution, discussed later, could be to add extra noise, to diminish the~relations between the~mines and their heights.
\end{itemize}

The test set consists of 3\:860 unlabeled observations. Approximately 25\% of them were used 
for evaluation on the~preliminary leaderboard, which was updated throughout the~contest when participants 
submitted their solutions. 
The~remaining observations were used for selection of the~best solutions at the~end of the~competition.

We should also note that the~observations in
the~test
set were randomly selected events rather than time series as provided within the training set.
More precisely, given a series consecutive observations, 
samples were uniformly drawn from them to form a test set. 
If two samples collected laid within the same window of 32 hours 
(for 24-hour long time series describing seismic activity plus 8 hour window for prediction), one
of them was dropped so as to assure that the samples were approximately independent. This procedure removes a 
significant amount of observations hence the size of the competition test set was relatively modest 
in comparison to the amount of training data available. This resulted in a very unreliable leaderboard
evaluation during the competition that was based on ca. 1\:000 observations. Therefore, we put great emphasis and
efforts to develop reliable evaluation methods given the available training data as discussed in the next section.

\subsection{Initial remarks}

When we approached the problem we quickly realized that the~main challenge was
to develop a~prediction model that generalizes well to new locations. 
Table~\ref{table:tse} presents the~warning frequencies per location in the~dataset. We observe that first of all, 
different locations vary considerably in terms of the~frequency of warnings. 
Secondly, the~sets of locations differ between  the~training
and test dataset. Additionally, the test set in the competition originated from future recordings
with respect to the training data available. This is the root of the problem. 
Hence a proposed model should be both location and time independent in the sense that 
it yields unbiased predictions for instances with no regard to their
origin and time they are collected. We also see that the~number 
of instances originating from different locations varies considerably. 
These preliminary observations 
should be carefully considered during model building and evaluation steps. We elaborate on this in the~next section.

\section{The solution}
In this section we describe in detail our solution to the~given problem. We discuss different 
sets of features that were proposed, various evaluation methods, models and their set up.

\subsection{Feature engineering} \label{sec:features}
In our experiments we created several feature sets for model training. 
For the~sake of simplicity and completeness, we describe them under consecutive headers
and denote as \textbf{FS}$_\mathbf{n}$ which stands for the~$n$-th
\textit{feature set} we proposed. These feature sets were 
developed independently by members of our team. Inevitably, there are significant overlaps between them.

\noindent\Name{FS}{1}: The~processing of the~data focused mainly on aggregation, 
aiming to reduce the~number of the~hourly measurements 
as the~majority of them were just zeros (for the~training set, about 66\% of all numbers were $0$). 
The feature extraction step ended up with 133 features, over
4 times less than the~original set.
From the~original features we kept:
\begin{itemize}
 \item all \textit{general} features;
 \item all seismic assessments converted to consecutive integers and their average;
 \item number of bumps (\texttt{count\_e*}) and their energies (\texttt{sum\_e*}) summed over all 24 hours,
together with mean energies resulting from division (if \texttt{count\_e*} was 0, 
then we were substituting the~result by 0);
 \item number of bursts and the~highest bump energies were just summed.
\end{itemize}
We also aggregated the~remaining time series related to most active geophones (8 time series), 
however this time we introduced some 
aggregations over subsets of hourly measures based on their relative importance. The~process is described below.

\begin{figure}[tbp]
\centering
\includegraphics[width=\hsize]{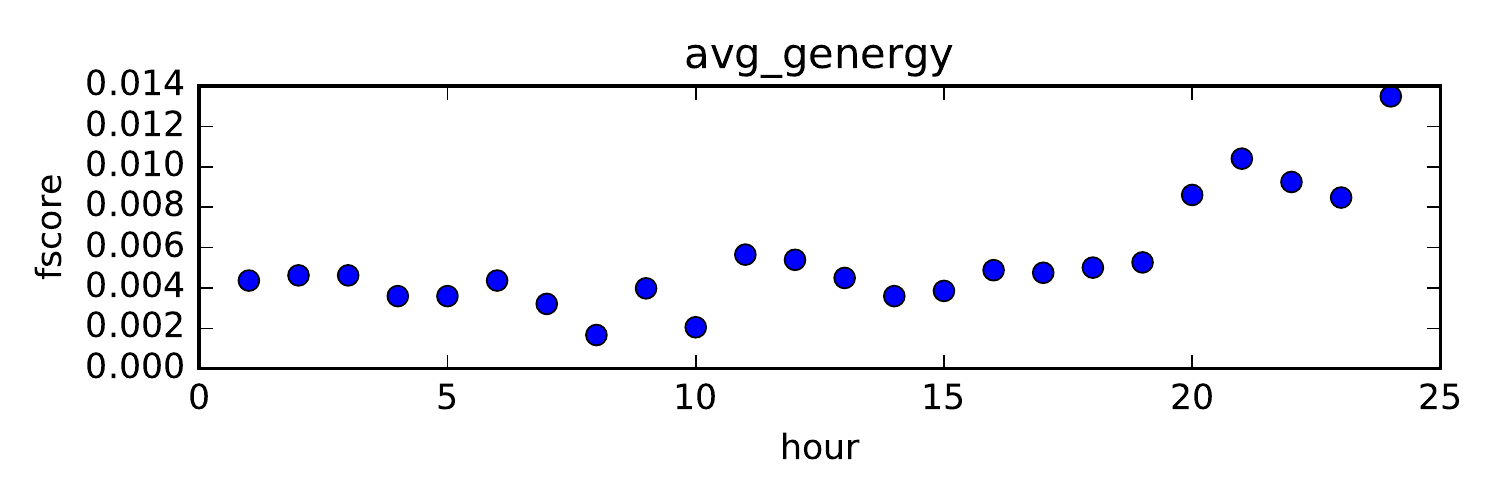}
\vspace{-0.8cm}
\caption{Importance of hourly measurements of \texttt{avg\_genergy}.}
\label{fig:fscore_avg_genergy}
\end{figure}

In order to assess the~impact of features we used a~functionality provided by the implementation of Gradient Boosting Trees available in the \textbf{XGBoost}~\cite{chen2016xgboost} package. The~library allows building a~tree classifier and assessing the~importance of particular features by providing the~number of times the~feature was used in a~split. The~more often a~feature is used, the~more separation gain it offers and therefore the~more important it is. We used an~XGBoost classifier with 150 trees (other parameters were default).
Fig.~\ref{fig:fscore_avg_genergy} presents an~example of such feature importance analysis for \texttt{avg\_genergy}. 
It seems that features are gaining importance towards the~end of the~time-series - it agrees with the~intuition that the~measurements closer to the~forecasting period are more informative. Therefore in this case, apart from the~entire time-series statistics, 
we are also interested in statistics based on the~last five hours (they stand out from the~preceding hours).
Also, we keep the~measurements from the~very last hour as a~separate feature. Having applied analogous analysis to the~above feature groups, 
we selectively compute statistics such as:
\begin{itemize}
 \item average and average over absolute values;
 \item standard deviation;
 \item max and max over absolute values;
 \item average over last $\gamma$ hours (were $\gamma$ varies from 1 to 6);
 \item standard deviation over last $\gamma$ hours;
 \item  slope of a~linear regression over last 5 hours with respect to time.
\end{itemize}

\begin{figure}[tbp]
\centering
\includegraphics[width=\hsize]{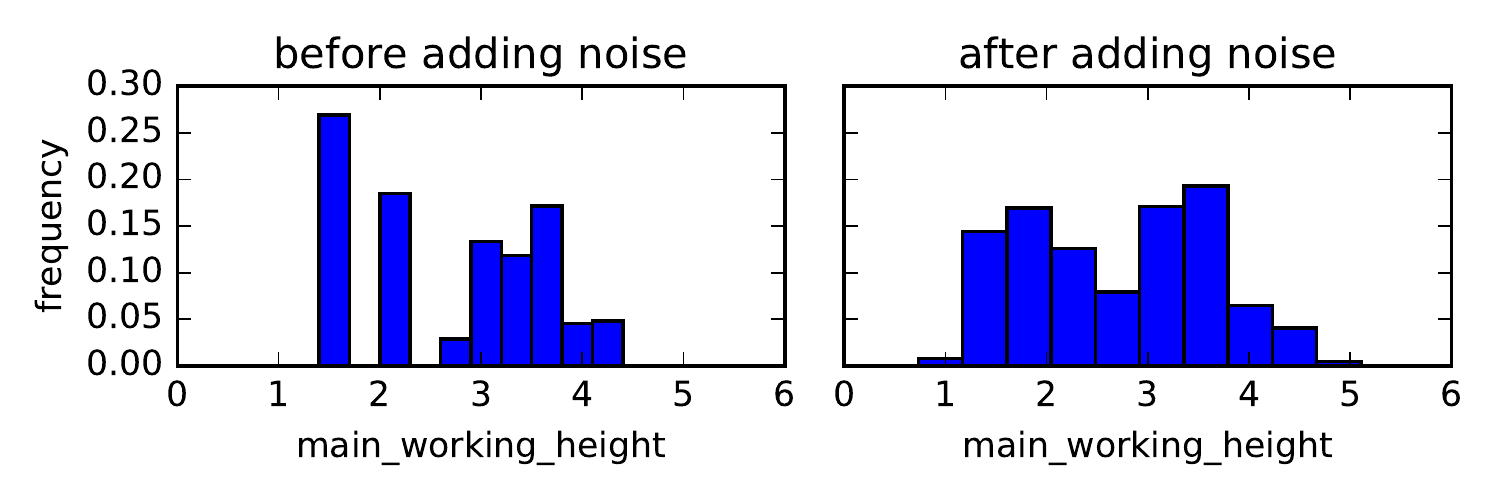}
\vspace{-0.8cm}
\caption{Distribution of \texttt{main\_working\_height}, before and after adding noise.}
\label{fig:hist_height}
\end{figure}

As mentioned above, competitors were also provided with the~metadata describing specific mine sites. Most of them were discarded. The~only metadata used here were the~\emph{main working height}, but only after adding Gaussian noise ($\sigma = 0.2$) resulting in more even distribution, see Fig.~\ref{fig:hist_height}. This step was performed 
to prevent a model from recognizing a particular location by its height.

In addition to the~above features, we produced a~vast set of more than 6\:000 interactions between them, 
i.e., pairwise products of features. 
This is obviously an~exhaustive number and we were not planning to use all of them. However, some interactions proved to be valuable. We applied an iterative process of selecting the most promising subsets of features and their interactions. 
We will come back to them when describing the~final model (Section \ref{sec:models} Model$_\mathbf{1}$).

\vspace{5pt}
\noindent\Name{FS}{2}:
In constructing this set, at first we decided to drop time series describing maximum statistics (\texttt{max*}) 
since they were highly correlated with corresponding average records.
For the~series \texttt{count\_e*}, \texttt{sum\_e*}, \texttt{number\_of\_rock\_bursts}, 
\texttt{number\_of\_destressing\_blasts}, \texttt{avg\_gactivity}, \texttt{avg\_genergy} 
we extracted the following features:
\begin{itemize}
    \item minimum,
    \item maximum,
    \item standard deviation,
    \item indicator variable, if there is a~non-zero value in the~series,
    \item hours elapsed from the~last non-zero observation.
\end{itemize}
Moreover, these statistics were derived over the~window of the~last 2, 4, 8 and 24 hours prior 
to the~forecasting period
in order to describe the~most recent data in greater detail.
These features were
appended to data with the~original time series that they were computed for. Additionally, for 
series \texttt{avg\_difference\_in\_gactivity} and \texttt{avg\_difference\_in\_genergy} 
maximal absolute value was derived
over the~last 2 and 24 hours. Finally, the~categorical variables were converted to binary features using 
one-hot encoding, i.e., for each possible value of a~categorical a~separate column was created
which indicates that a~given observation has this particular category. These operations produced a~feature set with a~total number of 700 features.

\vspace{5pt}
\noindent\Name{FS}{3}:
In this set of features, we first derived several new series based on the~original ones
\begin{enumerate}
    \item \texttt{log\_max\_avg\_diff\_genergy} which was derived as difference in \texttt{max\_genergy} and \texttt{avg\_genergy}, with application of logarithm hereafter. Analogous operation was performed for other series \texttt{avg} and \texttt{max\_difference\_in\_genergy} and a~corresponding series for \texttt{gactivity} 
    \item \texttt{log\_ave\_energy} series was produced by computing average energy based on \texttt{sum} and \texttt{count} series.
\end{enumerate}
In addition to features enumerated for \Name{FS}{2}, we derived the following statistics:
\begin{itemize}
    \item 0.25, 0.5 and 0.75-quantiles,
    \item number of times a~series increased in comparison to the~previous hour's recording,
    \item number of positive values in a~series,
    \item indicator variable, if there is a~non-zero value in the~series.
\end{itemize}
The statistics were computed on 4, 8 and 24 hours window. Furthermore, we computed the~coefficient, 
intercept and \texttt{R$^2$} statistic for a~fit of linear model of series \texttt{avg\_difference\_in\_gactivity}, 
\texttt{avg\_gactivity}, \texttt{log\_ave\_energy} to an~independent hourly temporal variable (1, 2, $\dots$, 24). 
Finally, we computed correlations between \texttt{avg\_difference\_in\_gactivity} and \texttt{avg\_difference\_in\_genergy}
as well as \texttt{avg\_gactivity} with \texttt{avg\_genergy}. After extracting features, constant features 
were dropped from the~feature set. 
Also, if there were features that were correlated over 0.99 (according to Pearson correlation coefficient), one of them was removed.
For categorical variables, they were one-hot encoded for logistic regression model or converted to integers 
(with higher risk categories being assigned a~higher integer) for tree-based models.

These steps produced a~training set of 426 features (for the~integer encoding of categorical features).

\vspace{5pt}
\noindent\Name{FS}{4}:
The feature set that was created with the~goal of being simple and 
as such leaving little room for overfitting. Out of the~basic (not time-based) features, \texttt{main\_working\_id} 
was dropped. Out of the~metadata, only \texttt{geological\_assessment}
was used. The~time-based features were ran through maximum
and standard deviation functions on 8-hour time periods with 4 hour increments, 
only the~features concerning quantities 
and maximums were used (features listing averages and sums were left out).

\subsection{Evaluation procedures}

Evaluation methodology is a~crucial part of creating a~successful application of a~model. 
Below we list  different validation techniques that we employed to assess the accuracy of a model. 
Here, the issue of overfitting a~model to particular locations and time-frames of samples is considered in detail.

$\newline$ \noindent
\textbf{$\mathbf{k}$-fold cross validation ($\mathbf{k}$-CV)}:
This is one of the~basic validation procedures. 
It is performed by assigning each example in the~training set randomly to one
of $k$ folds (in our application we used $k=10$ or $20$). Note that due 
to temporal alignment of instances in the~training data, 
this evaluation procedure tends to produce overly optimistic evaluation scores 
(we observed that during the contest by, e.g., large discrepancy in local evaluation and leaderboard scores).
This is because consecutive instances are likely to share the~same label. 
If some of them pertain to a~training fold and the~others to test fold, 
then a~classifier has a~relatively easy task to assign this instance to the~proper label.

$\newline$ \noindent
\textbf{Leave one location out (LOLO)}: This evaluation method was chosen to estimate the~model's performance on mining sites not
included in the~training data (see Table~\ref{table:tse}). It was supposed to promote models that overfit less and filter out 
those whose good performance was actually based on data leaks. 
We have decided to not use 
the three largest locations (with IDs 264, 373 and 437) for testing. These three locations 
constitute to a~large portion of the~total training data (48\%) and were not appearing in the~test set.
Locations that had no `warning' labels were also not used as validation data, as AUC could not be computed for them. 
This approach resulted in a 8-fold cross-validation
that gave much lower scores than the other ones (not even the best models could exceed 0.9 AUC), and the scores varied between folds (from as low as 0.6 to as high as 0.999) but it should not be percieved as a flaw --- it was intended behavior.

$\newline$ \noindent
\textbf{Train and test split \#1 (\Name{TrTs}{1})}:
This evaluation methodology was devised to reflect the~way the leaderboard was constructed. 
It is based on multiple train and test splits of the~data. It proceeds in two steps:
\begin{enumerate}
    \item 5 series are chosen at random and included in the~validation set,
    \item among series that have not been selected in 1) we include the first 70\% observations in the~training sample and the other 30\% in the~validation set.
\end{enumerate}
Moreover, in each of those 70\%-30\% splits, 32 observations between
the split point were 
removed to assert approximate independence between the~training and validation set (by introducing a gap of 32 hours between them). 
Again, data from locations with IDs 264, 373, 437 where included only in the~training set. 

In order to arrive at a reliable error estimate, this evaluation was repeated 25 times and consecutive 
measurements were averaged. With that many iterations we arrived at stable results for mean AUC value.

$\newline$ \noindent
\textbf{Train and test split \#2 (\Name{TrTs}{2})}:
The evaluation was based on multiple train and test splits (20 in the~final model) with some restrictions. By comparing the~\texttt{total\_seismic\_energy} (TSE) of mines (which turned out to be linearly correlated with the~frequency of appearances of warnings) we tried to make the~split, so the~TSE in the~inferred test sets resembled the~level of energies in the~private test set.

\begin{table}[ht]
\centering
\caption{Number of instances originating from different locations in the~training and test set along with averaged total seismic energies (TSE) and
frequency of warnings (not available for test set cases).}
\label{table:tse}
\begin{tabular}{crrrrr}
    \hline
      \multirow{3}{*}{Mine ID} &
      \multicolumn{3}{c}{Train set} &
      \multicolumn{2}{c}{Test set} \\
      & Instances & \specialcell{Mean \\ TSE [J]} & \specialcell{Warnings \\ Frequency} & Instances & \specialcell{Mean \\ TSE [J]} \\
    \hline
      373 & 31236 & 81002 & 1.1\% & - & - \\
      264 & 20533 & 7563 & 0.4\% & - & - \\
      725 & 14777 & 190232 & 9.4\% & 330 & 106741 \\
      777 & 13437 & 0 & 0.0\% & 330 & 29061 \\
      437 & 11682 & 4727 & 0.4\% & - & - \\
      541 & 6429 & 9397 & 0.9\% & 5 & 324 \\
      146 & 5591 & 678 & 0.1\% & 98 & 1 \\
      575 & 4891 & 9775 & 0.5\% & 253 & 7503 \\
      765 & 4578 & 136 & 0.0\% & 329 & 51265 \\
      149 & 4248 & 48357 & 7.3\% & 98 & 72749 \\
      155 & 3839 & 322021 & 17.2\% & 98 & 527229 \\
      583 & 3552 & 2595 & 0.2\% & 215 & 73302 \\
      479 & 2488 & 5548 & 0.0\% & 35 & 102 \\
      793 & 2346 & 0 & 0.0\% & 330 & 11547 \\
      607 & 2328 & 6027 & 0.0\% & 209 & 9470 \\
      599 & 1196 & 29932 & 1.9\% & 363 & 39962 \\
      171 & - & - & - & 49 & 33 \\
      470 & - & - & - & 258 & 10701 \\
      490 & - & - & - & 160 & 13698 \\
      508 & - & - & - & 58 & 32183 \\
      641 & - & - & - & 97 & 10672 \\
      689 & - & - & - & 83 & 63889 \\
      703 & - & - & - & 145 & 44031 \\
      799 & - & - & - & 317 & 8 \\
   \hline
\end{tabular}
\end{table}

\begin{figure}[tbp]
\centering
\includegraphics[width=\hsize]{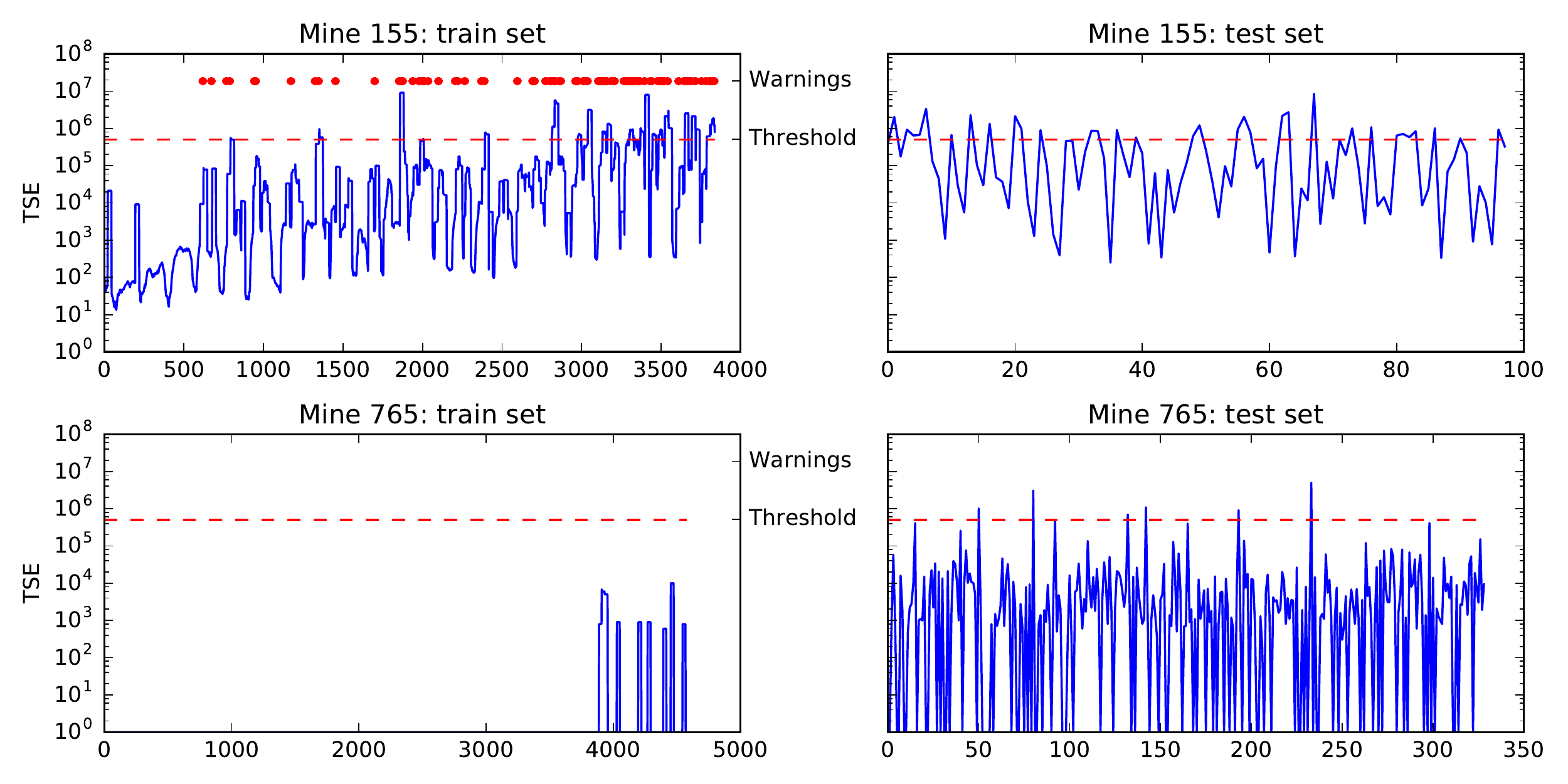}
\vspace{-0.8cm}
\caption{Total seismic energy over time.}
\label{fig:tse}
\end{figure}

Table~\ref{table:tse} presents averaged TSE for each mine grouped over train and test datasets, together with the~frequencies of warnings in the~training dataset. It is worth to point out significant discrepancies between the~activity levels of mines in both sets. For mine 765, the~activity in the~training set is mere 136 J, with no warnings. In the~test set, the~average activity is above 50 kJ, so there must have been several warnings emitted. A closer look reveals that there are some abnormalities in the~training set. Fig.~\ref{fig:tse} presents the~TSE of mines 155 and 765. While the~activity of the~former looks realistic, 765 is mute for majority of the~time, only to exhibit a few spikes towards the~end of the~time series. On the other hand, its activity in the~test set greatly increased. Some mines do not exhibit any activity in the~training set, i.e. TSE equals zero (mines 777, 793). This is one of the~reasons we have to avoid producing models that would be able to recognize the~mines, the~classifiers should generalize correctly from the~activity records, regardless any behavior specific to certain mines. Also, it poses a~problem - whether to consider the~suspicious mines during the~training or not. It is rather unusual for a~mine to have a~zero seismic activity and supposedly the~data in these periods might be corrupted.

The final train and test splits were based on the~above TSE analysis and were produced in the~following way:
\begin{enumerate}
    \item mines 777, 793 were excluded due to their suspicious lack of any activity in the~training set (although they represented a~significant amount of data);
    \item in every split, five randomly selected mines were left only for testing (to evaluate the~generalization properties of classifiers);
    \item from the~remaining sites we were taking 20\% of samples for testing. For mines 146 and 599 samples were drawn from the~beginning (due to corresponding energy levels in the~test set), for the~remaining - from the~end;
    \item in some cases (mines 373, 437) only samples where TSE were nonzero were taken into account.
\end{enumerate}
The process was repeated several times to obtain multiple train/test splits. The~final evaluation was based on the~average score over 20 splits.

\subsection{Model training and optimization} \label{sec:models}
There are several models that we employed in creating the~final solution to the~problem. We used the~implementation
of models available in Python's \textbf{scikit-learn} package for machine learning (ver. 0.17.1)~\cite{Sklearn01,Sklearn02}
and \textbf{XGBoost} package for tree boosting models (ver. 0.4)~\cite{chen2016xgboost}.
Throughout the~paper, for brevity, we use the~following abbreviations for the~model names: Linear Discriminant Analysis - LDA, 
Logistic Regression - LR, Extra Trees Classifier - ETC (all from \textbf{scikit-learn} library) 
and Extreme Gradient Boosting Classifier - XGB (from \textbf{XGBoost} library).

$\newline$ \noindent
\Name{Model}{1}:
The fist model was built using \Name{FS}{1} and TrTs$_2$ evaluation method.
Several models were considered, apart from XGB and ETC, also logistic regressions and neural networks, finally only 
the first two were used in the~final blend. They were performing particularly well in spite of rather large number of features.

First, we ran learning on all the~features and interactions we produced. Based on the~importance scores provided by XGB (described in Section \ref{sec:features} \Name{FS}{1}) we kept the first 982 interactions 
and all \emph{individual} features. Then, a~grid search returned sets of parameters scoring the~highest:

The optimal parameters for XGB model were (otherwise default):
\begin{itemize}
    \item \texttt{n\_estimators = 100}
    \item \texttt{max\_depth = 2}
    \item \texttt{learning\_rate = 0.08}
\end{itemize}
\noindent
The optimal parameters for ETC were (otherwise default):
\begin{itemize}
    \item \texttt{n\_estimators = 1000}
    \item \texttt{max\_depth = 7}
    \item \texttt{criterion = entropy}
\end{itemize}

It is worth to note, that trees, by their design, are relatively powerful in discovering interactions between features. However, in their case the~interactions are not discovered concurrently, but rather in a~multilevel manner, between consecutive splits. By explicitly using interactions as features, they can be made use of directly.

Having obtained well performing hyperparameters, we ran a~randomized search for best features' subsets. In each iteration we were randomly selecting from 20 to 40 individual features (out of 133) and additionally up to 10 interactions (out of 982).
We ran several thousands evaluations 
on XGB and several hundreds on ETC, tracking their validation scores.

The idea was to produce many models built only on subsets of features and to take advantage of assembling them which 
reduces variance of predictions and minimize the~risk of overfitting to anomalies in particular features. This is a~powerful method for increasing the performance of the model~\cite{ho1998random}.

The final blend was composed of:
\begin{itemize}
 \item single ETC of 10~000 trees using all 133 features and 20 best interactions;
 \item single XGB using the~same features;
 \item a~blend of 20 ETCs built on 20 best subsets of features; 
 \item a~blend of 20 XGBs built on 20 best subsets of features; 
\end{itemize}

The final submission scored 0.9199 on the~public leaderboard. The~score in the~final evaluation reached 0.9393 and turned out to be the best in the competition.

$\newline$ \noindent
\Name{Model}{2}:
The second model involved the~following classifiers: two linear models (LDA and LR) as well as the tree-based ETC model. 

The first part of the~solution was the LDA model trained on \Name{FS}{2} using $k$-CV evaluation procedure. The~regularization 
shrinkage parameter selection was done in an automated way (i.e., the~parameter \texttt{shrinkage} 
set to ``\texttt{auto}'') in \textbf{scikit-learn}'s LDA implementation. The~other models were LR and ETC
trained on \Name{FS}{3} using TrTs$_1$ evaluation method. 
The~parameter values were set using grid search. The optimal values for LR model were:
\begin{itemize}
    \item \texttt{penalty = l1}
    \item \texttt{C = 0.003}
\end{itemize}
\noindent
The optimal parameters for ETC were:
\begin{itemize}
    \item \texttt{n\_estimators = 1000} (number of trees)
    \item \texttt{max\_depth = 3}
    \item \texttt{max\_features = 200}
    \item \texttt{min\_samples\_split = 3}
    \item \texttt{class\_weight = 10} (for label `1').
\end{itemize}
The three models were blended by averaging their predictions with equal weights
to produce a~solution. Prior to averaging, the~model predictions were standardised so that 
their standard deviations would equal 1. This step aims to convert the~probabilities yielded 
by individual models to the~same scale. 
Note that the~mean values of predictions are irrelevant since AUC is invariant to
monotonic transformations of output, see Equation~\ref{AUC}. 
On the~competition test set, the~model yielded
0.9385 and 0.9340 of preliminary and final evaluation score, respectively.

$\newline$ \noindent
\Name{Model}{3}:
This model used only \Name{FS}{4} and was meant to be more universal than the~other models and 
thus was tuned on LOLO validation.
The algorithms used were ETC, XGB and logistic regression. For each algorithm, many sets of predictions were generated
(using the~top results from a~grid search). This model achieved $0.928$ and $0.933$ 
on preliminary and final evaluations, respectively.

Below we list the~best parameters found for each algorithm:\\
ETC
\begin{itemize}
 \item \texttt{min\_samples\_leaf = 5}
 \item \texttt{n\_estimators = 40\:000}
\end{itemize}
XGB
\begin{itemize}
 \item \texttt{subsample = 1.0}
 \item \texttt{num\_round = 200}
 \item \texttt{max\_depth = 10}
 \item \texttt{objective = binary:logistic}
 \item \texttt{base\_score = 0.05}
 \item \texttt{eta = 0.04}
 \item \texttt{colsample\_bytree = 0.8}
\end{itemize}
LTR
\begin{itemize}
 \item \texttt{solver = sag} (Stochastic Average Gradient, just for speed)
 \item \texttt{C = 1.0}
\end{itemize}

\subsection{Model ensemble}
We have decided to use sorted order position averaging
(as the AUC assessment method considers only the rank of predicted likelihoods and not the values)
of the three presented models' predictions with
the~final weights being 1, 3 and 2 for models 1, 2 and 3 respectively. The averaging was employed in order 
to leverage various approaches and come up with yet a better predictor for the given task. 
The weights for the ensemble were chosen basing on individual model's scores on 
the preliminary leaderboard. The ensemble produced a model scoring 0.933 and 0.938 
on preliminary and final leaderboard, respectively.
All in all, it turned out that model 1 
outperformed the full ensemble by a small margin (0.939 to 0.938). 
However, it might be caused by the relatively small test set size.

\subsection{Things we tried that did not work}
\begin{figure}[tbp]
\centering
\includegraphics[width=\hsize]{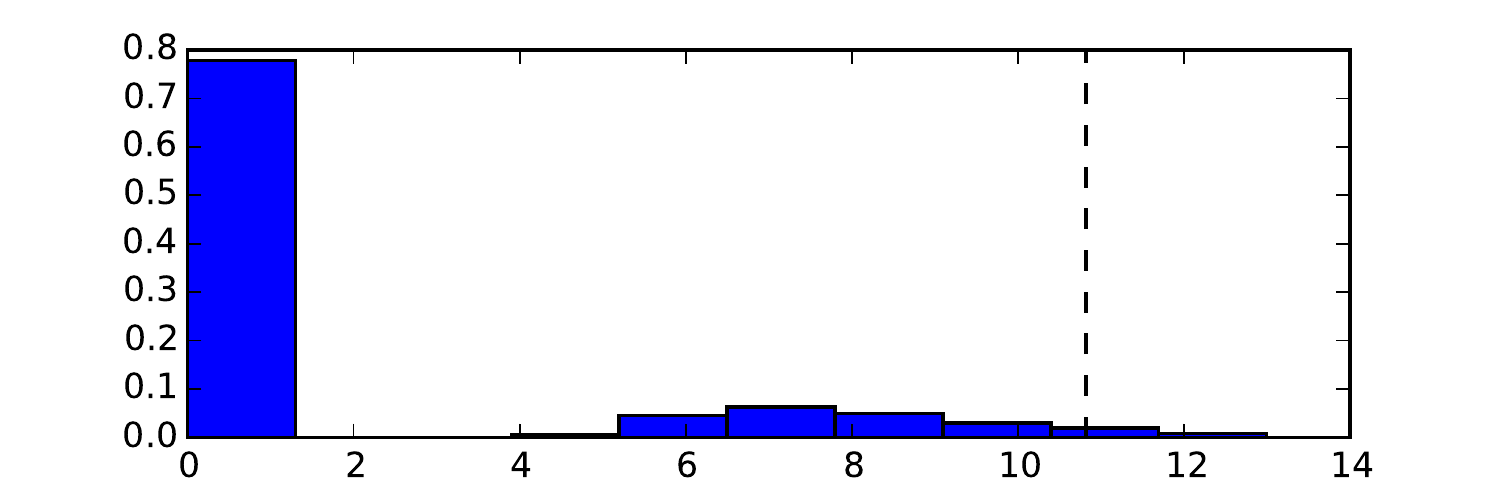}
\vspace{-0.8cm}
\caption{Histogram of the~total energy within 8 next hours (after with application of 
$\log(x+1)$ transformation). The~dashed line indicates the
warning level of $\log(5 \cdot 10^4 + 1)$ J.}
\label{fig:energy8h}
\end{figure}

Throughout the process of creating the most successful model we tested  a couple of ideas
that turned out not to improve our results.
First of all, we framed the~problem given in the~competition as a~regression task. 
Because the observations in the training data were given as time series, it was possible to retrieve the~energy level for the~next hours, 
see Fig.~\ref{fig:energy8h}. This allowed 
us to forecast energy levels within the~target window of 8 hours. Note that predictions from a~regression model may 
be directly evaluated using AUC accuracy measure as it can be considered as a~\textit{risk scoring} 
model for high values of energy. For the original classification problem, we tried to modify the~energy levels
and train the~models on an~enriched set of labels, e.g., we assumed 30 kJ (and a~couple of other values) 
as the~warning level and estimate the~model.

We also experimented with undersampling of training instances pertaining to class 0 so that 
the~proportion of 
`1' in the~training data increases. We also tried to reduce samples from locations 264, 373 and 437 
in the~training set by undersampling or assigning them a~lower sample weight (in, e.g., LR model).

However, in this particular application, our efforts were not successful 
as the~performance of the models (in terms of evaluation scores) 
was not improving.

\subsection{Model performance on the final test set}

\begin{figure}[tbp]
\centering
\includegraphics[width=\hsize]{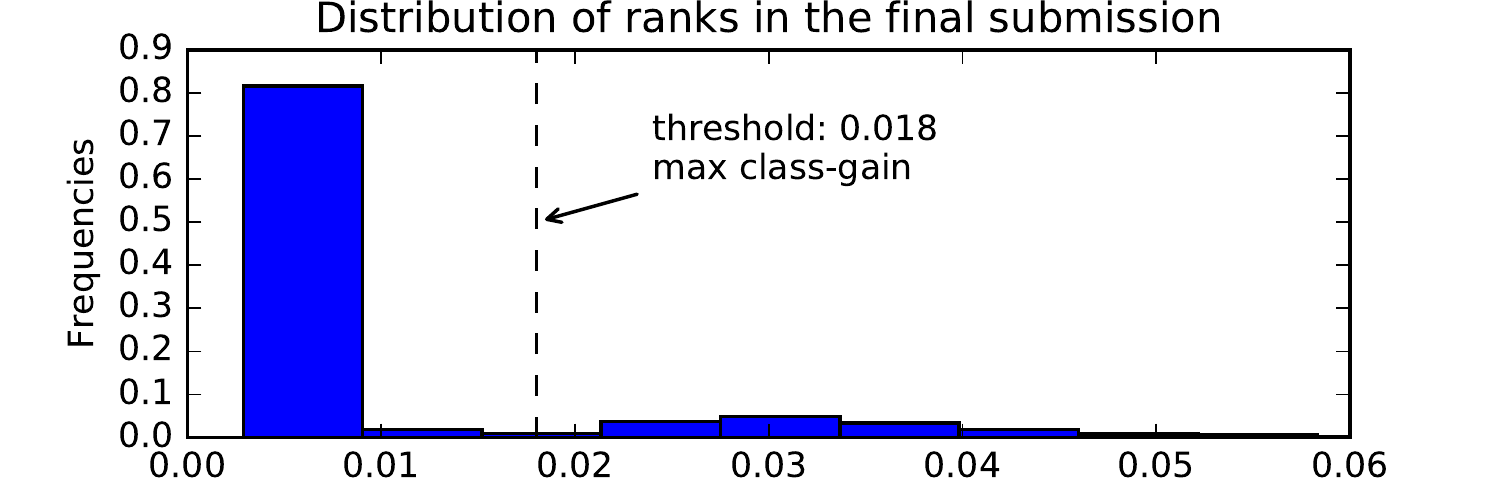}
\vspace{-0.8cm}
\caption{Histogram of likelihoods returned by the~winning model.}
\label{fig:hist_ranks}
\end{figure}

After the~competition we were provided with the~true labels used during the~final evaluation.
We were able to compute different metrics than AUC. The~winning model's predictions had a strongly skewed
distribution (Fig. \ref{fig:hist_ranks}), corresponding to total energies seen in Fig.~\ref{fig:energy8h}.
The~distribution has two modes, however of a~much lower mode related to predicted warnings - this is due to
imbalance of classes. Depending on the~threshold beyond which we consider predictions as warnings we can
derive the~confusion matrix:

\vspace{2mm}
\begin{tabular}{l|l|c|c|c}
\multicolumn{2}{c}{}&\multicolumn{2}{c}{True warning}&\\
\cline{3-4}
\multicolumn{2}{c|}{} & ~~~1~~~ & ~~~0~~~ \\
\cline{2-4}
\multirow{2}{*}{Predicted warning}
& ~~~1~~~ & $TP = 126$ & $FP = 284$ \\
\cline{2-4}
& ~~~0~~~ & $FN = 11$ & $TN = 2390$ \\
\cline{2-4}
\end{tabular}
\vspace{2mm}

\noindent
Based on the matrix we can compute several useful accuracy measures of the model:
\begin{align}
& \texttt{precision = TP / (TP + FP)} \\
& \texttt{sensitivity (recall) = TP / (TP + FN)} \\
& \texttt{specificity = TN / (TN + FP)} \\
& \texttt{F1} = \texttt{2} \cdot ~ \frac{\texttt{precision} \cdot \texttt{recall}}{\texttt{precision + recall}}\\
& \texttt{Class-gain = specificity + recall - 1}
\end{align}

\noindent
The threshold maximizing the~class-gain score is $0.018$ (see Fig.~\ref{fig:hist_ranks}) 
and yields the following accuracy on the final test set:

\vspace{2mm}
\begin{tabular}{ccccc}
    \hline
      precision & recall & specificity & F1 & class-gain \\
    \hline
      0.31 & 0.92 & 0.89 & 0.46 & 0.81 \\
   \hline
\end{tabular}
\vspace{2mm}

\begin{figure}[tbp]
\centering
\includegraphics[width=\hsize]{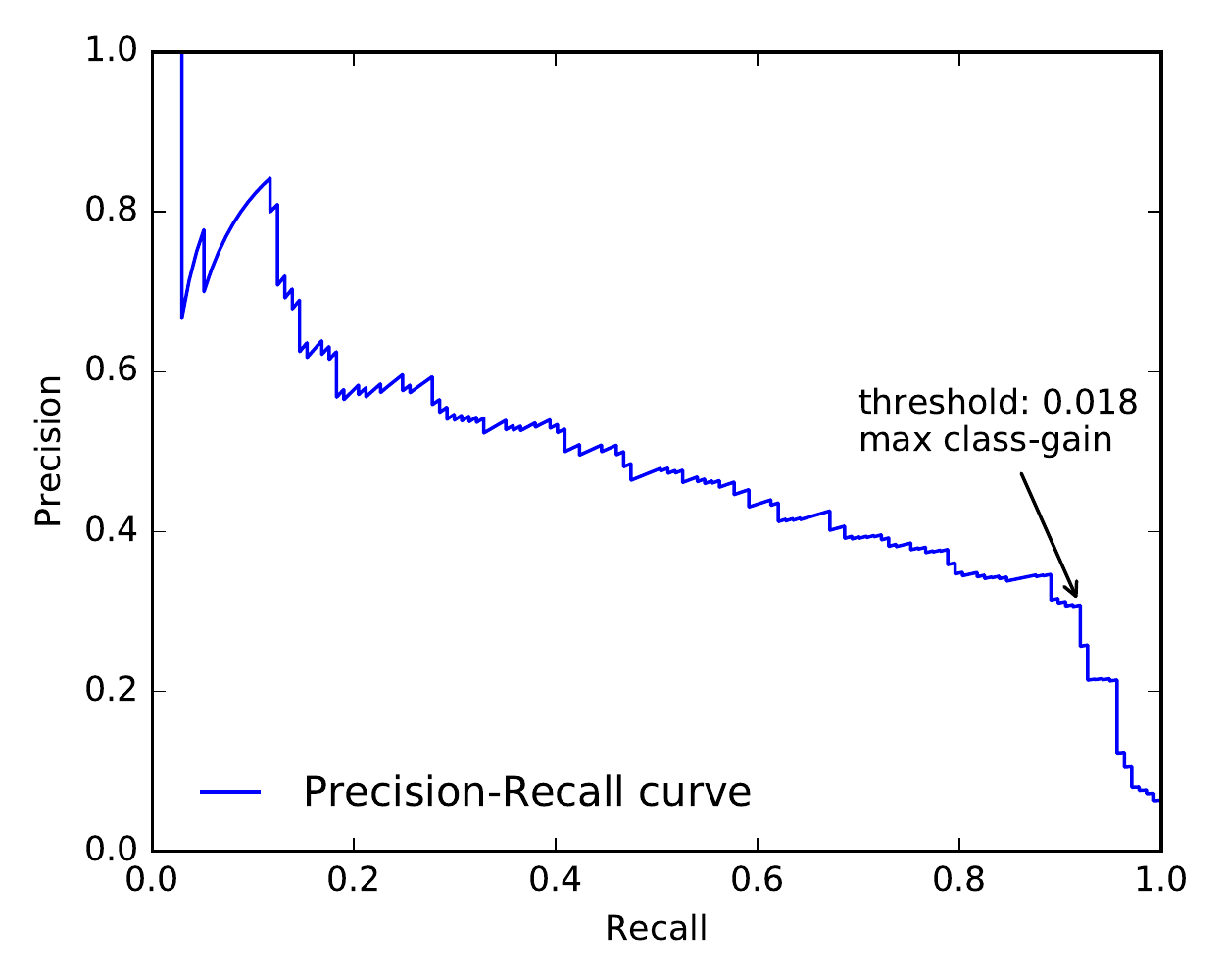}
\vspace{-0.8cm}
\caption{Precision-recall curve.}
\label{fig:prc}
\end{figure}

The entire precision-recall curve can be seen in Fig.~\ref{fig:prc}.

In order to assess our results we looked for research addressing similar problems as the~one considered here. 
In~\cite{sikora2011induction} we found an~approach to solve the~same problem, however the~results are not directly comparable since they are based on different datasets. Our results prove to outperform
results reported there for all presented classifiers 
(the highest class-gain reported is 0.75).
Also, in the cited paper there was no inter-coal-mine validation, while the~models described in our work were cross-validated on separate coal mines. Therefore, the models proposed
are designed to generalize well and should be applicable also to working sites with no historical data available.

\section{Summary}

Given that the~dataset originates from working mine sites, with the~entire measurement infrastructure already 
installed, we hope that the~approach presented in this paper could be implemented and serve 
as a~valuable tool for alerting about dangerous seismic events early. This hopefully should result in preventing possible accidents
which pose a threat to employees and generate losses from damaged coal mine infrastructure and machinery.

Even though the~models presented here have outperformed the~other models in the competition, we recommend
they be ensembled with other high-scoring models, because properly combined efforts of multiple participants
are expected to yield better results than individual solutions.

Lastly, we would like to thank the~organizers for the~opportunity to solve
a~real-life problem and the contestants for creating such a~competitive environment.

\balance

\bibliographystyle{IEEEtran}
\bibliography{references}

\begin{thebibliography}{1}
\providecommand{\url}[1]{#1}
\csname url@samestyle\endcsname
\providecommand{\newblock}{\relax}
\providecommand{\bibinfo}[2]{#2}
\providecommand{\BIBentrySTDinterwordspacing}{\spaceskip=0pt\relax}
\providecommand{\BIBentryALTinterwordstretchfactor}{4}
\providecommand{\BIBentryALTinterwordspacing}{\spaceskip=\fontdimen2\font plus
\BIBentryALTinterwordstretchfactor\fontdimen3\font minus
  \fontdimen4\font\relax}
\providecommand{\BIBforeignlanguage}[2]{{%
\expandafter\ifx\csname l@#1\endcsname\relax
\typeout{** WARNING: IEEEtran.bst: No hyphenation pattern has been}%
\typeout{** loaded for the language `#1'. Using the pattern for}%
\typeout{** the default language instead.}%
\else
\language=\csname l@#1\endcsname
\fi
#2}}
\providecommand{\BIBdecl}{\relax}
\BIBdecl

\bibitem{WUG2015}
\BIBentryALTinterwordspacing
\relax{Wyższy Urząd Górniczy}, ``Wypadkowość w górnictwie od 1 stycznia
  2015 do 31 grudnia 2015,'' 2015, in Polish, last accessed 18 April 2016.
  [Online]. Available:
  \url{http://www.wug.gov.pl/bhp/Statystyki_archiwalne_2015}
\BIBentrySTDinterwordspacing

\bibitem{Zagorecki2015}
\BIBentryALTinterwordspacing
A.~Zagorecki, \emph{Rough Sets, Fuzzy Sets, Data Mining, and Granular
  Computing: 15th International Conference, RSFDGrC 2015, Tianjin, China,
  November 20-23, 2015, Proceedings}.\hskip 1em plus 0.5em minus 0.4em\relax
  Cham: Springer International Publishing, 2015, ch. Prediction of Methane
  Outbreaks in Coal Mines from Multivariate Time Series Using Random Forest,
  pp. 494--500. [Online]. Available:
  \url{http://dx.doi.org/10.1007/978-3-319-25783-9_44}
\BIBentrySTDinterwordspacing

\bibitem{Janusz2016}
A.~Janusz, M.~Sikora, {\L}.~Wr{\'o}bel, and D.~\'Sl\k{e}zak, ``{Predicting
  Dangerous Seismic Events: AAIA16 Data Mining Challenge},'' in
  \emph{Proceedings of FedCSIS 2016}.\hskip 1em plus 0.5em minus 0.4em\relax
  IEEE, 2016, in print September 2016.

\bibitem{sikora2011induction}
M.~Sikora, ``Induction and pruning of classification rules for prediction of
  microseismic hazards in coal mines,'' \emph{Expert Systems with
  Applications}, vol.~38, no.~6, pp. 6748--6758, 2011.

\bibitem{KnowledgePit2016}
\BIBentryALTinterwordspacing
\relax{Knowledge Pit, a host platform for data challenges}, 2016, last accessed
  18 April 2016. [Online]. Available: \url{https://knowledgepit.fedcsis.org/}
\BIBentrySTDinterwordspacing

\bibitem{chen2016xgboost}
T.~Chen and C.~Guestrin, ``Xgboost: A scalable tree boosting system,''
  \emph{arXiv preprint arXiv:1603.02754}, 2016.

\bibitem{Sklearn01}
F.~Pedregosa, G.~Varoquaux, A.~Gramfort, V.~Michel, B.~Thirion, O.~Grisel,
  M.~Blondel, P.~Prettenhofer, R.~Weiss, V.~Dubourg, J.~Vanderplas, A.~Passos,
  D.~Cournapeau, M.~Brucher, M.~Perrot, and E.~Duchesnay, ``Scikit-learn:
  Machine learning in {P}ython,'' \emph{Journal of Machine Learning Research},
  vol.~12, pp. 2825--2830, 2011.

\bibitem{Sklearn02}
L.~Buitinck, G.~Louppe, M.~Blondel, F.~Pedregosa, A.~Mueller, O.~Grisel,
  V.~Niculae, P.~Prettenhofer, A.~Gramfort, J.~Grobler, R.~Layton,
  J.~VanderPlas, A.~Joly, B.~Holt, and G.~Varoquaux, ``{API} design for machine
  learning software: experiences from the scikit-learn project,'' in \emph{ECML
  PKDD Workshop: Languages for Data Mining and Machine Learning}, 2013, pp.
  108--122.

\bibitem{ho1998random}
T.~K. Ho, ``The random subspace method for constructing decision forests,''
  \emph{Pattern Analysis and Machine Intelligence, IEEE Transactions on},
  vol.~20, no.~8, pp. 832--844, 1998.

\end{thebibliography}

\end{document}